# Addressing the Pitfalls of Image-Based Structural Health Monitoring: A Focus on False Positives, False Negatives, and Base Rate Bias

Vagelis Plevris[1*]

ORCiD ID: 0000-0002-7377-781X

[1] Department of Civil and Environmental Engineering, Qatar University
P.O. Box: 2713, Doha, Qatar
e-mail: vplevris@qu.edu.qa

*Corresponding author

**Abstract.** This study explores the limitations of image-based structural health monitoring (SHM) techniques in detecting structural damage. Leveraging machine learning and computer vision, image-based SHM offers a scalable and efficient alternative to manual inspections. However, its reliability is impacted by challenges such as false positives, false negatives, and environmental variability, particularly in low base rate damage scenarios. The Base Rate Bias plays a significant role, as low probabilities of actual damage often lead to misinterpretation of positive results. This study uses both Bayesian analysis and a frequentist approach to evaluate the precision of damage detection systems, revealing that even highly accurate models can yield misleading results when the occurrence of damage is rare. Strategies for mitigating these limitations are discussed, including hybrid systems that combine multiple data sources, human-in-the-loop approaches for critical assessments, and improving the quality of training data. These findings provide essential insights into the practical applicability of image-based SHM techniques, highlighting both their potential and their limitations for real-world infrastructure monitoring.

## 1    Background and Motivation

Structural health monitoring (SHM) of civil infrastructure plays a crucial role in sustainable development. SHM involves the in situ, non-destructive measurement of the operating and loading conditions, as well as the critical responses of a structure. Damage-sensitive features are extracted from this data and statistically analyzed to detect the presence, location, and severity of structural damage. SHM also helps determine the current health condition of a structure, estimate its remaining useful life, and guide engineers and inspectors in making informed decisions regarding maintenance, rehabilitation, or replacement of infrastructure [1].

Traditionally, structural inspections relied heavily on manual evaluations performed by engineers or technicians who would visually assess the state of a structure. While these inspections remain a crucial part of infrastructure maintenance, they are limited by subjective judgment, accessibility issues, and the vast number of structures that require regular monitoring. In recent years, the use of image-based classification methods has seen a significant rise in SHM and infrastructure management [2]. These methods, powered by advancements in artificial intelligence (AI) [3, 4], machine learning (ML) [5] and computer vision techniques [6], offer a way to supplement or even replace manual inspections by analyzing large volumes of image data captured by drones, cameras, or sensors [7] and they are increasingly being applied to detect damage in critical structures such as bridges [8], buildings, and tunnels [9]. By automating the process of damage detection, these technologies have the potential to revolutionize traditional inspection methods, which are time-consuming, labor-intensive, and prone to human error.

At the heart of these image-based techniques are algorithms designed to classify or segment images to detect potential signs of damage, such as cracks [10], corrosion, deformation, spalling [11], and others [12]. Convolutional neural networks (CNNs), deep learning (DL) models, and other artificial intelligence (AI) approaches are commonly used for this purpose. These models can be trained on large datasets of labeled images to recognize patterns that are indicative of structural damage, thus automating the detection process with high speed and accuracy.

One of the key motivations for adopting image-based techniques in SHM is their scalability and efficiency. Drones equipped with high-resolution cameras can survey large structures in a fraction of the time it would take for manual inspections [13]. Furthermore, AI models can analyze these images in real time, providing almost immediate feedback on the condition of the structure [2]. This rapid detection capability is especially critical in emergency situations, such as after an earthquake or a severe storm, where quick assessments are necessary to ensure public safety.

Additionally, image-based methods can capture minute details that might be missed by the human eye, especially in hard-to-reach areas or over extended periods where damage progression is subtle. The use of such techniques enables continuous monitoring and early detection of problems, potentially preventing costly and dangerous structural failures. Payawal et al. [14] conducted a systematic review of image-based SHM techniques. Their study highlights that image-based SHM represents a technological breakthrough aimed at addressing existing uncertainties in civil engineering and construction. However, several challenges still need to be overcome. Another state-of-the-art review on AI-assisted visual inspection systems has been carried out by Mishra and Lourenço, this time focusing on cultural heritage structures [15].

However, despite these promising developments, the reliability of image-based classification methods in terms of damage detection in real-world applications is not without challenges. Issues such as false positives (where damage is incorrectly identified, when it does not exist) and false negatives (where the system fails to identify existing damage) remain a concern. Furthermore, while these technologies excel in controlled environments or with high-quality data, their effectiveness in diverse and complex real-world settings, where lighting, angles, and environmental factors vary, is less clear.

Given the potential inaccuracies and the low occurrence rate of actual damage in most structures, the significance of a positive result from these models must be carefully scrutinized. This becomes particularly crucial when considering the safety risks associated with undetected damage, as well as the financial burden of false positives, which can lead to unnecessary repairs and wasted resources. In response to these challenges, this paper aims to examine the limitations of image-based damage detection techniques, focusing on the effects of false positives, false negatives, and the Base Rate Fallacy [16]. By critically evaluating the practical effectiveness of these methods, the study seeks to determine whether they can reliably support the maintenance of structural integrity or if their limitations undermine their utility in certain contexts. Additionally, this study proposes several strategies to mitigate these limitations and enhance the reliability of image-based SHM systems.

## 2  Overview of Image-Based Techniques for Damage Detection

Image-based techniques have gained significant traction in the field of SHM, driven by advances in ML, DL, and computer vision technologies. Automated inspection systems equipped with drones or stationary cameras are commonly employed to capture high-resolution images of hard-to-reach areas in structures like bridges, dams, and high-rise buildings [8]. These images are then processed through ML models, which analyze the data for signs of damage without the need for manual intervention. The combination of drones, high-resolution imagery, and DL algorithms is transforming traditional inspection processes by automating tasks that previously required significant labor and time.

At the forefront of these techniques are Convolutional Neural Networks (CNNs), a specialized type of DL model that excels at recognizing patterns and features in images [17]. CNNs are particularly useful for detecting surface-level damage such as cracks, corrosion, or spalling in structural components [18]. By training CNNs on large datasets of labeled images, these models can learn to identify damage patterns with impressive accuracy [19].

Other DL methods, including Recurrent Neural Networks (RNNs) and hybrid architectures, are also being explored to account for more complex structural behaviors and damage patterns over time [20]. Computer vision techniques, which involve the use of algorithms to analyze and interpret visual data from cameras or sensors, have been widely adopted for detecting surface-level deformations or anomalies in structures [21]. These technologies often rely on advanced algorithms for image segmentation, edge detection, and pattern recognition to identify potential damage.

## 2.1 Advantages of Image-Based Methods

The primary advantage of image-based techniques in damage detection is their ability to automate and scale the inspection process [18]. Traditional manual inspections are labor-intensive, time-consuming, and prone to human error, especially when dealing with large or complex structures. Image-based methods, on the other hand, can quickly analyze vast amounts of visual data, reducing the need for on-site personnel and providing faster assessments.

Additionally, these techniques allow for **continuous monitoring**. By using cameras integrated with real-time data analysis, structures can be continuously inspected without the need for scheduled manual assessments. This real-time capability is particularly valuable in the early detection of damage, enabling preventative maintenance before small issues escalate into larger structural problems [22]. Image-based techniques can also be complemented by additional data, such as information from sensors and other instruments, to enhance accuracy and reliability.

Another key benefit is the ability to **access difficult-to-reach areas**. Drones equipped with high-resolution cameras can inspect areas that are dangerous or otherwise inaccessible for human inspectors, such as the underside of bridges or tall skyscrapers [8]. The use of drones also enables more frequent inspections at a fraction of the cost, contributing to the overall efficiency of the monitoring process.

Furthermore, the **scalability** of these techniques makes them ideal for monitoring large infrastructure networks. From a city's network of bridges to a country's roadways, image-based methods can be deployed on a large scale, providing comprehensive coverage and reducing the time required to detect potential issues.

## 2.2 Challenges in Image-Based Damage Detection

While image-based classification techniques have shown great potential for automating damage detection in structures, they face several key challenges, primarily related to the accuracy and reliability of the results. A known limitation of these methods has to do with their **dependence on high-quality data**. The performance of DL models, including CNNs, is highly reliant on the quality of the images used for training and analysis. Images with poor resolution, or those affected by noise or environmental factors, can significantly degrade the model's ability to correctly classify damage [23]. Furthermore, these methods are often tailored to surface-level damage, making it difficult to detect internal structural problems such as subsurface cracks or material fatigue, which might not be visible through imagery alone.

Additionally, the **variability in environmental conditions**—such as lighting, weather, and perspective—can introduce noise or distortions in the images, reducing the effectiveness of damage detection algorithms [24]. For example, a crack detected in a sunny, clear image may go undetected in an image taken under cloudy or shadowy conditions. This variability presents challenges in maintaining consistent accuracy across different inspection scenarios.

In addition, **training DL models** requires large and diverse datasets of labeled images [25]. In many cases, collecting and labeling enough high-quality images of damaged and undamaged structures can be a time-consuming and resource-intensive process. Furthermore, the rarity of actual structural damage in many datasets (low base rate) complicates the training process, making it difficult for models to learn to differentiate between true damage and benign anomalies.

Another major concern arises from the presence of false positives and false negatives—two types of classification errors that can significantly impact the decision-making process in SHM. **False positives** occur when the image-based system incorrectly identifies damage in a structure where none exists. This type of error (**Type I error**) can lead to unnecessary inspections, repairs, and resource allocation. Conversely, **false negatives** represent an even greater challenge in structural damage detection, as they occur when the system fails to detect actual damage. This type of error (**Type II error**) can have severe safety implications, as undetected damage may worsen over time, leading to structural failures or even catastrophic incidents.

## 3  Understanding False Positives and False Negatives in Damage Detection

Both false positives and false negatives highlight the trade-offs inherent in using image-based classification techniques. While these systems offer scalability and efficiency, the risks associated with classification errors cannot be ignored. Even small error rates can have outsized impacts when dealing with safety-critical infrastructure. As such, engineers and decision-makers must consider not only the accuracy of these models but also the significance and consequences of the errors they may produce.

A **confusion matrix** is a performance evaluation tool used in classification problems to summarize how well a ML model or classification algorithm has performed [26]. It is a table that displays the number of true positive (*TP*), true negative (*TN*), false positive (*FP*), and false negative (*FN*) predictions, providing insights into the types of errors the model makes. The matrix helps assess the model's accuracy, precision, recall, and other performance metrics. Each cell in the confusion matrix corresponds to the actual versus predicted outcomes, making it a valuable tool for evaluating classification algorithms where multiple types of predictions are involved.

Figure 1 presents a confusion matrix for the case of damage detection. The confusion matrix helps reveal how often the system makes each type of error, which is crucial for understanding the trade-offs between identifying more damage and avoiding false alarms. The figure provides also the basic formulas for the calculation of useful statistical quantities, such as the *Accuracy*, *Precision*, and *Recall* of the system [27].

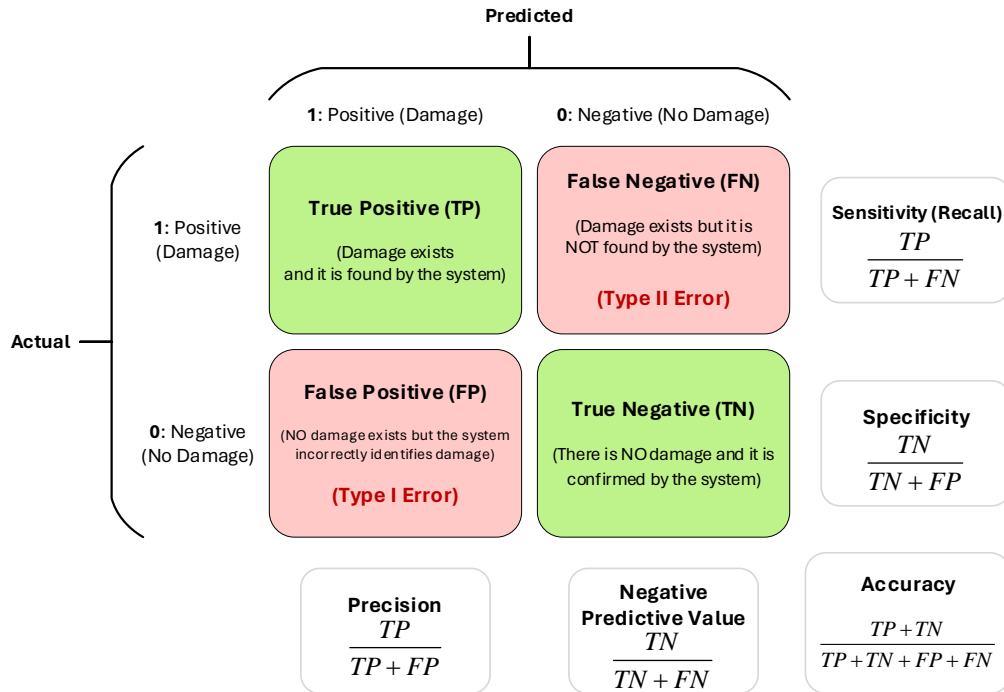

Figure 1. Confusion matrix for a damage identification problem.

In practice, increasing the precision of a model often results in a decrease in recall, and vice versa. The *F1-score* captures the balance between these two metrics in a single value, which can be expressed as:

$$F1 = \frac{2}{\frac{1}{Recall} + \frac{1}{Precision}} = \frac{2 \cdot Recall \cdot Precision}{Recall + Precision} \qquad (1)$$

The F1-score is the harmonic mean of precision and recall, providing a comprehensive measure that reflects the balance between these metrics. It reaches its maximum value when precision is equal to recall. Both false positives (corresponding to Type I Errors) and false negatives (corresponding to Type II Errors) present unique challenges in the context of SHM, and understanding their implications is critical for engineers and decision-makers relying on image-based methods. The presence of such errors can significantly undermine trust in these methods, particularly when used for safety-critical infrastructure. In large-scale SHM programs, where hundreds or thousands of structures are routinely inspected, even a small percentage of these errors can have considerable consequences.

While **false positives** may seem less critical than false negatives, they can lead to a significant misallocation of resources. When a system incorrectly identifies damage, maintenance teams may be dispatched to inspect or repair undamaged structures, resulting in unnecessary costs and labor. In a worst-case scenario, if the frequency of false positives becomes too high, decision-makers might lose confidence in the system, leading to underuse or disregard of the technology altogether. This lack of trust can stall the adoption of automated methods, pushing engineers back to manual inspections, which are slower and more costly.

**False negatives** are arguably more problematic because they represent a failure to detect actual damage. This type of error is particularly dangerous in safety-critical structures such as bridges, tunnels, or large buildings, where undetected damage could compromise structural integrity over time. If damage goes unnoticed, it may progress to a point where repairs are no longer possible, increasing the risk of catastrophic failure. In public infrastructure, the consequences of false negatives can be dire, leading to accidents, loss of life, and significant legal and financial liabilities for asset managers and government bodies.

These inaccuracies complicate decision-making for engineers. They must continuously balance the need for fast, efficient damage detection with the inherent risks of relying on automated systems prone to classification errors. Engineers may find themselves second-guessing the results of the system, needing to introduce additional layers of manual verification, which defeats the purpose of automation in the first place.

### 3.1 Base Rate Fallacy and its role in SHM

The *Base Rate Fallacy* [16], also known as *base rate bias* and *base rate neglect* [28], is a cognitive bias where people tend to ignore or underweight the base rate (i.e., the general probability of an event occurring) in favor of specific information or evidence presented, leading to erroneous conclusions. This fallacy occurs in situations where the base rate of an event—such as a disease, accident, or failure—is relatively low, but the likelihood of a positive result (such as a medical test or detection method) is mistakenly interpreted without adequately considering the initial low probability of the event [29].

The Base Rate Fallacy can often arise in various fields, such as medical diagnostics [30, 31], criminal justice [32], and financial risk analysis. In the medical field, for example, even a highly accurate test for a rare disease might yield a disproportionately high number of false positives because the disease itself occurs so infrequently [33]. Despite the high accuracy of the test, the low occurrence of the disease means that the majority of positive results may not correspond to actual cases of the disease. The fallacy occurs when individuals focus too heavily on the test result and neglect to consider the overall rarity of the condition .

The fallacy can also manifest in public health scenarios, particularly during outbreaks like the COVID-19 pandemic. A common misconception involves the effectiveness of vaccines in highly vaccinated populations [34]. Some people may conclude that vaccines are ineffective simply because the majority of infections occur among vaccinated individuals. However, this reasoning neglects the base rate of vaccination in the population, leading to misleading interpretations. In highly vaccinated populations, it is expected that vaccinated individuals will represent a significant portion of infection cases simply because they constitute the vast majority of the population [34]. However, this observation alone does not imply that the vaccine is ineffective—it highlights the importance of evaluating outcomes in relation to the base rates of the population rather than focusing narrowly on case counts.

In general, this fallacy is particularly prevalent when evaluating ML models or any detection system that operates in environments where the events being detected occur at a very low rate. The problem is exacerbated when people intuitively expect that a positive result from a seemingly accurate system must indicate a high probability of the event occurring, without accounting for the low base rate. This will be highlighted in section 4 of this study through the use of a practical example.

### 3.2 Conditional Probabilities and Bayes' Theorem

Conditional probability refers to the probability of an event *A* occurring given that another event *B* has already taken place. It is expressed as *P(A|B)*, meaning the probability of *A* happening, assuming *B* has occurred. This concept is often described as "*A* given *B*". The probability of *A* depends on the prior occurrence of B and is calculated using Bayes' theorem [35], which helps estimate the likelihood of an outcome based on new information.

Bayes' rule provides a framework for updating the probability of a hypothesis (*A*) when relevant evidence (*B*) becomes available [36]. It states that the conditional probability of event *A*, given event *B*, is equal to the likelihood of event *B* occurring given *A*, multiplied by the prior probability of *A*, and then divided by the probability of *B*. The formula is as follows:

$$P(A|B) = \frac{P(B|A) \cdot P(A)}{P(B)} \quad (2)$$

Where:

- *P*(*A*) is the **prior probability** of *A*, which represents the likelihood of *A* before considering any new evidence.
- *P*(*B*) is the **marginal probability** of B, representing the overall likelihood of observing event *B*.
- *P*(*A*|*B*) is the **posterior probability**, or the probability of *A* occurring given that *B* has happened.
- *P*(*B*|*A*) is the **likelihood**, or the probability of observing event *B* if *A* is true.

In cases where events A and B are independent, it is *P*(*A*|*B*) = *P*(*A*) and *P*(*B*|*A*) = *P*(*B*), meaning the occurrence of one event does not influence the probability of the other.

Bayes' Theorem plays a critical role in a wide range of fields, offering a powerful tool for reasoning about probabilities and updating beliefs in the presence of new information. Its significance lies in its ability to combine prior knowledge (or assumptions) with fresh evidence to refine the probability of an event. This approach is particularly valuable when dealing with uncertain or dynamic environments where data evolves over time.

One of the major strengths of Bayes' Theorem is its flexibility in handling complex problems involving uncertainty. It allows us to incorporate existing knowledge (prior probabilities) and adjust our understanding based on new observations, enabling more informed decision-making. This process of updating beliefs is iterative—each new piece of evidence refines our prior knowledge, resulting in a more accurate posterior probability.

In the broader context, Bayes' Theorem finds applications across many disciplines, such as Medical Diagnostics, ML and AI [36], Risk Analysis and Decision-Making, Forensics and Legal Reasoning [37], Search and Rescue Operations [38, 39], Marketing and Consumer Behavior [40], and others. In all these applications, the ability of Bayes' Theorem to update probabilities based on real-time data is invaluable. It provides a structured and quantitative approach to dealing with uncertainty, making it essential in scenarios where decision-making relies on balancing probabilities with new, often incomplete, information. This process of continuously refining predictions or hypotheses is one of the key reasons why Bayes' Theorem remains a cornerstone in fields that require precise, data-driven insights.

## 4 Numerical Example in SHM

In this section, we will examine the efficiency of an image-based SHM system with high accuracy, while also considering the base rate of damage in a city. We will demonstrate that, even if the system exhibits theoretically high performance in detecting damage—characterized by a high true positive rate—it is still extremely likely to trigger false alarms in most examined cases if the base rate of damage is relatively low. To understand this phenomenon, we apply Bayes' theorem to calculate the probability that a positive diagnosis by the system is indeed correct. We also investigate the relationship between key performance parameters of the system and the base rate of damage, and propose strategies to mitigate the challenges associated with low base rate environments.

We consider a city with thousands of buildings of varying sizes and ages. In this scenario, only a small fraction of these buildings—approximately 1 in every 1,000—has a structural defect. For simplicity, each building is classified as either "intact" or "damaged" in a binary classification, without any intermediate states. To ensure the safety and integrity of its infrastructure, the city has implemented an advanced, autonomous SHM system. This system uses drones equipped with high-resolution cameras that continuously scan and capture thousands of images of each building, providing a comprehensive visual record of the structures.

The SHM system is fully automated: after collecting images, it uploads the data to the cloud, where digital imaging procedures analyze the photos. Using advanced DL and AI algorithms, the system classifies whether damage is present or not. The system is highly efficient. According to its documented specifications:

- It has a **98% success rate** in detecting damage when it actually exists, meaning that in 98 out of 100 cases with real damage, the system successfully identifies that damage exists. In other words, the system misses damage in only 2% of the cases with actual damage present.

- In addition, like all systems, it occasionally produces false positives, identifying damage where none exists, at a **rate of 5%**. In other words, in 95% of cases with no damage the system will also find no damage.

Now, we will examine what happens when the system detects damage in one of the city's buildings. Based on its high success rate according to its manufacturer, many people and even experts might instinctively believe that a "positive" result from such an advanced and theoretically accurate system would lead to a high probability that the building is actually damaged. However, when we factor in the base rate of damage, the reality becomes far less intuitive.

To understand this, we break down the problem using the following information:

- **Base rate of damage ($b$)**: Only 1 in 1,000 buildings ($b$=0.1%) has actual structural damage.

- **True positive rate ($TPR$)**: If there is damage, the system detects it 98% of the time and fails to detect it 2% of the time ($TPR$=98%). This means that the False Negative Rate is $FNR$=2%.

- **False positive rate ($FPR$)**: The system mistakenly detects damage in 5% of undamaged buildings and it identifies correctly that there is no damage in 95% of the cases of undamaged buildings ($FPR$=5% and True Negative Rate $TNR$=95%)

Now, we would like to determine the probability that a building is actually damaged, given that the system has flagged it as damaged (i.e., the system gives a positive result) and taking into account the base rate of damage in the city. Since 1 in 1,000 buildings (0.1%) has actual structural damage, then for the general population of buildings:

- $P(damaged) = 0.001$
- $P(intact) = 1 - P(damaged) = 0.999$

Our system appears to be quite efficient, with a 98% accuracy (Recall value) in detecting damage when it actually exists. Let $T$ denote a positive test result of the system (the system predicts structural damage). Thus, we have that:

$$P(T \mid damaged) = 0.98 = TPR \qquad (3)$$

The system occasionally produces false positives, identifying damage where none exists, with a false positive rate of 5%.

$$P(T \mid intact) = 0.05 = FPR \qquad (4)$$

In the above, *P*(*T* | *damaged*) represents the conditional probability that the test is positive, given that the building is damaged, while P(*T* | *intact*) represents the conditional probability that the test is positive, given that the building is intact (not damaged). The test mistakenly indicates damage in 5% of cases when the building is intact, so this probability is 0.05.

In this problem, we try to calculate the conditional probability P(*damaged* | *T*), i.e. the probability that a building is actually damaged, given a positive test result from the system. According to Bayes' theorem, it is:

$$P(damaged \mid T) = \frac{P(T \mid damaged) \cdot P(damaged)}{P(T)} \qquad (5)$$

To do the above calculation, we also need to find the probability, i.e. the probability of a positive test result *P*(*T*). This is given by:

$$P(T) = P(T \mid damaged) \cdot P(damaged) + P(T \mid intact) \cdot P(intact) \qquad (6)$$

Which gives us

$$P(T) = 0.98 \cdot 0.001 + 0.05 \cdot 0.999 = 0.05093 = 5.093\% \qquad (7)$$

As a result,

$$P(damaged \mid T) = \frac{0.98 \cdot 0.001}{0.05093} = \frac{98}{5093} \approx 0.01924 = 1.924\% \qquad (8)$$

This surprising result means that the probability of the building being actually damaged, given a positive test result by the system, is less than 2%, which is counterintuitive considering the system's theoretical high accuracy. Given the high success rate that the manufacturer of the system reports (98%), one would expect that the probability of a building being damaged based on a positive test result would be very high. On the contrary, this probability for the particular example is less than 2%, which is a very low probability and practically gives no value to any decision maker.

We can reach the same conclusion using a frequentist approach without directly relyingon the Bayes' Theorem, by reasoning as follows:

- Suppose we inspect 100,000 buildings in the city.

- Out of these, 100 buildings have damage (1 per thousand), while the remaining 99,900 buildings are intact (undamaged).

- Since the system falsely indicates damage in 5% of cases where there is no actual damage, 5% of the 99,900 intact buildings, or 4,995 buildings, are incorrectly flagged as damaged.

- Additionally, the system correctly identifies 98% of the 100 damaged buildings, meaning 98 buildings are accurately flagged as damaged, while 2 damaged buildings are missed.

- Therefore, the total number of buildings reported as damaged by the system is 4,995 + 98 = 5,093 buildings.

- Thus, the probability that a building flagged as "damaged" by the system is actually damaged is 98/5,093 ≈ 0.01924, or approximately 1.924%.

In this example, if one presents the confusion matrix using the *TPR*, *FNR*, *FPR*, *TNR* rates, without taking into account the number of cases and the base rate of damage in the city, one can obtain the misleading version of the confusion matrix presented in Table 1. It has to be noted that the rows of the matrix presented in Table 1 have to sum up to 100%, but that is not the case with the columns. Using this matrix, one may expect that the precision of the system is very high in any practical situation.

Table 1. Confusion matrix of the hypothetical SHM system, using rates (percentage values).

|  | **1. Positive (Damage)** | **0. Negative (No damage)** | **SUM** |
|---|---|---|---|
| **1. Positive (Damage)** | *TPR*=98% | *FNR*=2% | **100%** |
| **0. Negative (No damage)** | *FPR*=5% | *TNR*=95% | **100%** |

However, if the base rate of damage in the city is taken into account (0.1% in this example) and we consider a specific number of cases (100,000 buildings in this example), we obtain the correct confusion matrix of Table 2 for our example.

Table 2. Confusion matrix of the hypothetical SHM system,
using the base rate of damage (0.1%) and 100,000 examined buildings in total.

|  | **1. Positive (Damage)** | **0. Negative (No damage)** | **SUM** |
|---|---|---|---|
| **1. Positive (Damage)** | *TP* = 98 | *FN* = 2 | **100** |
| **0. Negative (No damage)** | *FP* = 4995 | *TN* = 94905 | **99,900** |
|  | **5093** | **94907** | **100,000** |

Then for the system presented in Table 2, the performance metrics can be calculated using the equations presented in Figure 1 and Eq. (1), as follows:

- *Accuracy* = 0.95003 = 95.00%
- *Precision* = 0.019242097 = 1.92%
- *Recall* = 0.98 = 98.00%
- *F1* = 0.037743116

We see that using the confusion matrix of Table 2, we obtain the correct precision value of 1.92% which is exactly the conditional probability P(*damaged* | *T*), that was previously calculated using Bayes' theorem and the frequentist approach. The precision metric expresses the probability that a building is actually damaged, given a positive test result from the system. A value of 1.92% means that less than 2 buildings out of 100 flagged as "damaged" are actually damaged.

## 5   Parametric investigation

We consider the following basic quantities in a parametric investigation:

- *TPR*: The true positive rate (98% in the previous example)
- *FPR*: The false positive rate (5% in the previous example)
- *b*: The base rate of damage (0.1% in the previous example)

- $N$: The number of examined cases (100,000 in the previous example)

The first two of the above parameters, *TPR* and *FPR*, are characteristics of the SHM system, while the third one, *b*, is a characteristic of the city being examined, while *N* is the number of buildings examined (sample size). In this case, the formulas giving the *TP*, *TN*, *FP*, and *FN* values (cases) depend on the sample size *N* and they are given by:

$$TP = N \cdot b \cdot TPR \tag{9}$$

$$FN = N \cdot b \cdot (1-TPR) \tag{10}$$

$$FP = N \cdot FPR \cdot (1-b) \tag{11}$$

$$TN = N \cdot (1-b) \cdot (1-FPR) \tag{12}$$

On the other hand, the performance metrics of the system do not depend on the sample size *N*, and they are given by the formulas:

$$Accuracy = 1 - FPR \cdot (1-b) - b \cdot (1-TPR) \tag{13}$$

$$Precision = \frac{b \cdot TPR}{FPR \cdot (1-b) + b \cdot TPR} \tag{14}$$

$$Recall = TPR \tag{15}$$

$$F1 = \frac{2b \cdot TPR}{FPR \cdot (1-b) + b \cdot (1+TPR)} \tag{16}$$

The above proposed formulas for the *Accuracy*, *Precision*, *Recall* and *F1-score* should be used in cases where the *TPR*, *FPR* rates are known, and also the base rate of damage is either known or it can be efficiently approximated using known information. By observing Eqs. (13)-(16) we see that all performance metrics (with the only exception of the *Recall* value) depend strongly on the base rate of damage, *b*. The base rate of damage in the city must be taken into account in order to access the significance of a positive test result.

## 5.1 The special case of TPR=100%

In the special case where the True Positive Rate (*TPR*) is 100% (i.e., the False Negative Rate *FNR* is 0), the system achieves perfect detection of damage—meaning that whenever there is damage, the system identifies it every single time. However, false positives can still occur, as the False Positive Rate (FPR) is not necessarily zero, indicating that the system may incorrectly identify damage where none exists. This is a simpler, special case of the general case examined in the previous section, and it can be used to extract useful results.

With *TPR* = 100%, the performance metrics of the system can be simplified using the following formulas:

$$Accuracy = 1 - FPR \cdot (1-b) \tag{17}$$

$$Precision = \frac{b}{FPR \cdot (1-b) + b} \tag{18}$$

$$Recall = 1 \tag{19}$$

Addressing the Pitfalls of Image-Based Structural Health Monitoring:
A Focus on False Positives, False Negatives, and Base Rate Bias$$F1 = \frac{2b}{FPR \cdot (1-b) + 2b} \qquad (20)$$

If we consider the previous example, keeping the False Positive Rate (*FPR*) at 5% (i.e., True Negative Rate (*TNR*) = 95%) but increasing the *TPR* to 100% (from the previous 98%), the performance metrics can be recalculated as follows:

- *Accuracy* = 0.95005 = 95.01% (previously 95.00%)

- *Precision* = 0.0196271 = 1.96% (previously 1.92%)

- *Recall* = 1 = 100% (previously 98.00%)

- *F1* = 0.038498556 (previously 0.037743116)

Even with *TPR* = 100%, we notice that the Precision of the system only slightly increases, from 1.92% to 1.96%. This means that a positive test result still implies only a 1.96% probability that actual damage is present. Figure 2 graphically depicts Eq. (18), i.e. the values of *Precision* as a function of *b* and *FPR* (for the case *TPR* = 100%).

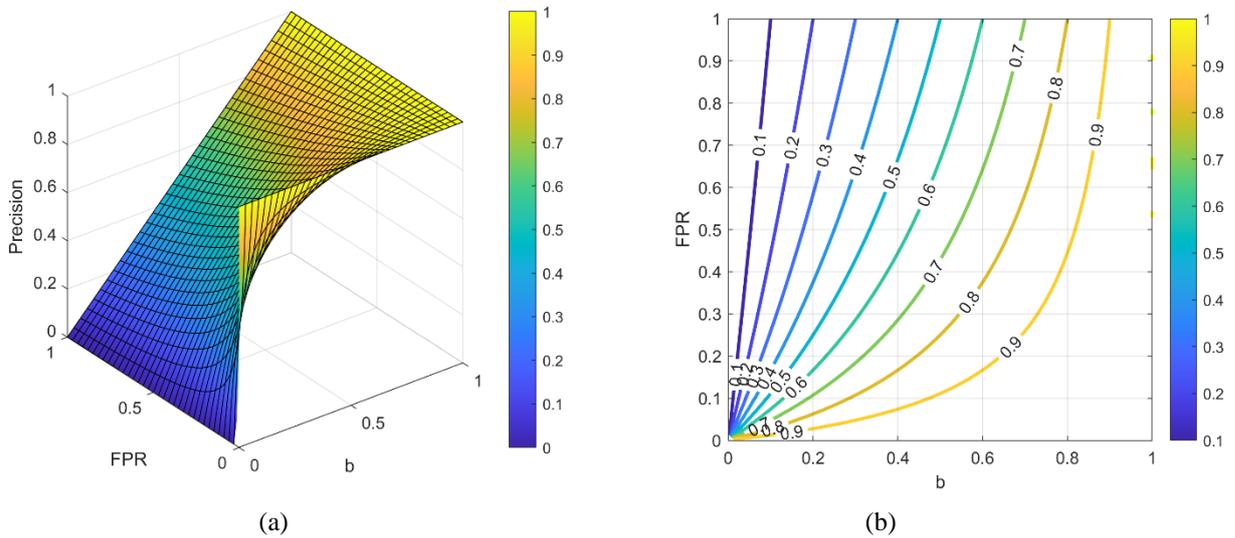

(a)          (b)

Figure 2. Precision as a function of *FPR* and *b* (for the case *TPR*=1): (a) Surface plot, (b) Contour plot.

Figure 3 focuses on the lower left part of Figure 2, i.e. on values of *b* and *FPR* up to 0.10 (or 10%). We see that for low levels of the base damage rate, extremely small values of *FPR* are required by the system to achieve satisfactory values of *Precision*.

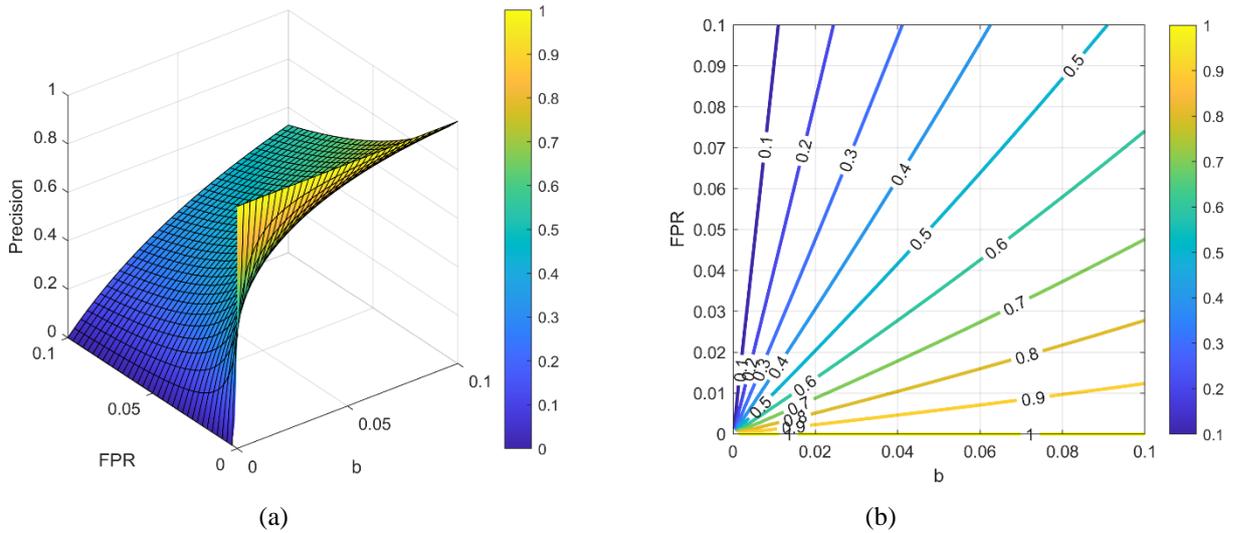

Figure 3. Precision as a function of *FPR* and *b* (for the case *TPR*=1), zoomed in:
(a) Surface plot, (b) Contour plot.

We see that, for instance, to achieve *Precision* of 90% given a value of *FPR*=10%, the base rate of damage would have to be as high as b=47.4% which is too large as a base damage rate for any normal city. Or in another example, by focusing on *FPR*, for 90% *Precision* given that *b*=5%, the needed false positive rate would have to be as low as *FPR*=0.58%. In other words, for the case of 5% base damage rate in a city, the false positive rate should be lower than 0.58% in order to achieve precision higher than 90% for the system. Based on Eq. (18), the equation that provides the needed value of *FPR* for given values or *b*, *Precision* (for *TPR*=100%) is:

$$FPR = \frac{b}{Precision} \cdot \frac{1-Precision}{1-b} \qquad (21)$$

In the above equation, *Precision* $\neq 0$ and $b \neq 1$. We notice that for $b = 0$, then $FPR = 0$ for any value of the *Precision*, but this is a theoretical case where no damage exists in the city (base damage is zero), so in fact there is no point in applying the system. On the other hand, in the case of $b = Precision$, we obtain FPR=1 (or 100%) no matter the precision, which means that that for a precision value equal to the base rate of damage, there are no special requirements for FPR.

## 6  Evaluation of Significance: Do Image-Based Techniques Hold Value?

As image-based damage detection techniques become more prevalent in SHM, it is crucial to evaluate whether these methods truly hold practical value, particularly in light of the challenges posed by false positives and false negatives. While these systems offer scalability, automation, and the ability to monitor structures continuously, engineers must carefully assess when to trust a positive result and how to improve the reliability of these methods. The trade-offs between economic costs and safety risks are critical considerations that will determine the overall utility of image-based techniques in real-world applications.

One of the key challenges in evaluating image-based classification systems is determining when a positive result—indicating potential damage—can be trusted. Engineers must account for the fact that even highly accurate systems can produce false positives, especially when the base rate of actual damage is low. Blindly acting on every positive result can lead to unnecessary inspections, repairs, and operational disruptions.

To assess the value of a positive result, engineers can implement several strategies:

- **Thresholds and Confidence Scores**: Many ML systems provide not only a binary classification (damaged or undamaged) but also a confidence score that indicates the model's certainty about its prediction. Engineers can establish a threshold for confidence scores, acting on positive results only when the confidence level exceeds a certain value. For instance, if the model predicts damage with 95% confidence, this could warrant further investigation, while lower-confidence predictions might trigger additional verification steps.

- **Risk-Based Decision Making**: Engineers can prioritize responses to positive results based on the risk associated with the specific structure. For critical infrastructure—such as bridges or tunnels with high safety risks—a conservative approach may be taken, acting on positive results even at lower confidence thresholds. Conversely, for less critical structures, engineers may require stronger evidence before initiating costly maintenance procedures.

- **Secondary Validation Steps**: Before acting on a positive result, additional validation steps can be implemented. This might include a follow-up inspection using another detection method, such as ultrasonic testing, vibration analysis, or manual inspection, to confirm or rule out the presence of damage. By combining multiple sources of evidence, engineers can reduce the likelihood of acting on false positives, ensuring that resources are allocated efficiently.

To enhance the reliability of image-based damage detection systems and reduce the rates of false positives and false negatives, several approaches can be employed:

- **Hybrid Approaches**: One of the most effective ways to improve the reliability of damage detection is by integrating image-based techniques with other SHM methods. For example, combining visual data with sensor-based monitoring, such as vibration or acoustic sensors, can provide a more comprehensive view of a structure's health. While image-based methods excel at detecting surface-level damage, sensors can detect internal issues like material fatigue or subsurface cracks, complementing the visual data.

- **Human-in-the-Loop Systems**: Incorporating human oversight into the damage detection process can help reduce classification errors. In a human-in-the-loop system, initial damage detections from the ML model are reviewed by an expert engineer before any actions are taken. This approach leverages the strengths of automation while retaining the accuracy of human judgment in ambiguous or high-risk cases. Engineers can validate or override the system's predictions, ensuring that only the most reliable results are acted upon.

- **Improving Data Quality and Model Training**: The performance of image-based systems is highly dependent on the quality of the data used to train the models. Improving the dataset by incorporating more diverse and higher-quality images, including a wide range of damage types and environmental conditions, can significantly enhance the model's ability to differentiate between damaged and undamaged structures. Additionally, using data augmentation techniques—such as generating synthetic images of damaged structures—can help the model generalize better to real-world scenarios.

- **Adaptive Algorithms**: Another promising approach is the development of adaptive algorithms that can adjust their detection thresholds based on real-time data. These algorithms could, for instance, adjust their sensitivity based on the structural history, environmental conditions, or feedback from other SHM systems, reducing the likelihood of both false positives and false negatives.

In addition, in evaluating the value of image-based damage detection systems, engineers must weigh the economic costs associated with false positives against the safety risks posed by false negatives. In many cases, the trade-offs between economic costs and safety risks will depend on the specific application and the criticality of the structure being monitored. For safety-critical infrastructure, it may be prudent to adopt conservative detection thresholds and hybrid validation systems to minimize the risk of false negatives. For less critical applications, a more lenient approach may be taken, optimizing for cost-effectiveness by tolerating a certain level of false positives.

## 7 Conclusions

In this paper, we explored the limitations of image-based techniques for damage detection in SHM and examined how these limitations affect their practical significance. While advancements in ML and AI have brought significant potential for automating the inspection of structures, several challenges still pose barriers to the effective deployment of these methods in real-world applications. Chief among these challenges are the issues of false positives, false negatives, and the Base Rate Fallacy, all of which can critically undermine the reliability of image-based damage detection systems.

False positives—where damage is mistakenly identified in structures that are intact—can lead to unnecessary maintenance, driving up operational costs and overwhelming maintenance teams with false alarms. The financial and logistical burden of acting on false positives reduces the overall efficiency of automated SHM systems, especially when applied to large infrastructure networks. Conversely, false negatives, where the system fails to detect actual damage, present a far more dangerous scenario. Undetected damage can compromise the safety and integrity of critical structures, such as bridges, tunnels, and high-rise buildings. This type of error is particularly concerning for public safety, as it can lead to structural failures with potentially catastrophic consequences. Therefore, it is crucial to strike a balance between minimizing both types of errors to ensure that SHM systems are reliable enough to support informed decision-making.

A key aspect of the study is the role of the Base Rate Fallacy, which occurs when the low probability of structural damage is not adequately considered during the evaluation of positive results from damage detection systems. Even with highly accurate models, the rarity of actual damage in most structures can result in a low probability that a positive result truly indicates damage. This counterintuitive outcome highlights the importance of considering base rates when interpreting predictions from automated systems. Failure to do so can lead to misguided actions, as seen in many other cases where base rates were ignored.

To address these limitations and improve the reliability of image-based SHM systems, this paper proposes several strategies. First, hybrid monitoring systems that combine image-based techniques with complementary methods, such as acoustic or vibration-based monitoring, can provide a more comprehensive understanding of a structure's health. These techniques can detect both surface-level and internal damage, improving overall system accuracy. Second, incorporating human-in-the-loop approaches allows expert engineers to review automated classifications, reducing the risk of both false positives and false negatives. Finally, implementing risk-based decision frameworks can help prioritize maintenance efforts by focusing on safety-critical structures, ensuring that resources are used efficiently and effectively.

Future research should focus on further developing these hybrid systems and refining ML models to better account for the low base rates of damage typical in most SHM applications. Integrating additional data sources—such as sensor-based monitoring or historical maintenance records—into the training of ML models could enhance their ability to detect subtle or rare damage types, particularly in complex environments. Another important direction for future work is improving model robustness to environmental factors like lighting, weather conditions, and image quality, which can significantly affect damage detection accuracy. Research efforts should also focus on adaptive algorithms that can dynamically adjust detection thresholds based on real-time data, helping to mitigate the effects of false positives and negatives.

In conclusion, while image-based techniques offer scalability and efficiency in the realm of structural health monitoring, their current limitations necessitate a cautious approach to their adoption. Engineers and decision-makers must combine these technologies with additional validation methods, while also accounting for statistical biases, to make informed, data-driven decisions. By doing so, automated SHM systems can be harnessed to contribute more meaningfully to the maintenance and safety of critical infrastructure.

**Declarations**

**Funding**: No funding was received to assist with the preparation of this manuscript.

**Competing interests**: The authors have no competing interests to declare that are relevant to the content of this article.

**Data availability statement**: Data sets generated during the current study are available from the corresponding author on reasonable request.

## References

[1] Wang, G. and J. Ke, *Literature Review on the Structural Health Monitoring (SHM) of Sustainable Civil Infrastructure: An Analysis of Influencing Factors in the Implementation.* Buildings, 2024. **14**(2): p. 402 DOI: https://doi.org/10.3390/buildings14020402.
[2] Kim, J.-W., H.-W. Choi, S.-K. Kim, and W.S. Na, *Review of Image-Processing-Based Technology for Structural Health Monitoring of Civil Infrastructures.* Journal of Imaging, 2024. **10**(4): p. 93 DOI: https://doi.org/10.3390/jimaging10040093.
[3] Lagaros, N.D. and V. Plevris, *Artificial Intelligence (AI) Applied in Civil Engineering.* Applied Sciences, 2022. **12**(15) DOI: https://doi.org/10.3390/app12157595.
[4] Lu, X., V. Plevris, G. Tsiatas, and D. De Domenico, *Editorial: Artificial Intelligence-Powered Methodologies and Applications in Earthquake and Structural Engineering.* Frontiers in Built Environment, 2022. **8** DOI: https://doi.org/10.3389/fbuil.2022.876077.
[5] Plevris, V., A. Ahmad, and N.D. Lagaros, eds. *Artificial Intelligence and Machine Learning Techniques for Civil Engineering*. 2023, IGI Global. DOI https://doi.org/10.4018/978-1-6684-5643-9.
[6] Archana, R. and P.S.E. Jeevaraj, *Deep learning models for digital image processing: a review.* Artificial Intelligence Review, 2024. **57**(1): p. 11 DOI: https://doi.org/10.1007/s10462-023-10631-z.
[7] Spencer, B.F., V. Hoskere, and Y. Narazaki, *Advances in Computer Vision-Based Civil Infrastructure Inspection and Monitoring.* Engineering, 2019. **5**(2): p. 199-222 DOI: https://doi.org/https://doi.org/10.1016/j.eng.2018.11.030.
[8] Mandirola, M., et al., *Use of UAS for damage inspection and assessment of bridge infrastructures.* International Journal of Disaster Risk Reduction, 2022. **72**: p. 102824 DOI: https://doi.org/10.1016/j.ijdrr.2022.102824.
[9] Cha, Y.-J., R. Ali, J. Lewis, and O. Büyüköztürk, *Deep learning-based structural health monitoring.* Automation in Construction, 2024. **161**: p. 105328 DOI: https://doi.org/10.1016/j.autcon.2024.105328.

[10] Qayyum, W., R. Ehtisham, V. Plevris, J. Mir, and A. Ahmad, *Classification of Wall Defects for Maintenance Purposes using Image Processing*, in *9th ECCOMAS Thematic Conference on Computational Methods in Structural Dynamics and Earthquake Engineering (COMPDYN 2023)*. 2023: Athens, Greece. p. 2529-2539. DOI: https://doi.org/10.7712/120123.10580.21466.

[11] Dawood, T., Z. Zhu, and T. Zayed, *Machine vision-based model for spalling detection and quantification in subway networks.* Automation in Construction, 2017. **81**: p. 149-160 DOI: https://doi.org/10.1016/j.autcon.2017.06.008.

[12] Ehtisham, R., W. Qayyum, V. Plevris, J. Mir, and A. Ahmad, *Classification and Computing the Defected Area of Knots in Wooden Structures using Image Processing and CNN*, in *5th ECCOMAS Thematic Conference on Evolutionary and Deterministic Methods for Design, Optimization and Control (EUROGEN 2023)*. 2023: Chania, Crete, Greece. p. 10-21. DOI: https://doi.org/10.7712/140123.10187.18992.

[13] Akbar, M.A., U. Qidwai, and M.R. Jahanshahi, *An evaluation of image-based structural health monitoring using integrated unmanned aerial vehicle platform.* Structural Control and Health Monitoring, 2019. **26**(1): p. e2276 DOI: https://doi.org/10.1002/stc.2276.

[14] Payawal, J.M.G. and D.-K. Kim, *Image-Based Structural Health Monitoring: A Systematic Review.* Applied Sciences, 2023. **13**(2): p. 968 DOI: https://doi.org/10.3390/app13020968.

[15] Mishra, M. and P.B. Lourenço, *Artificial intelligence-assisted visual inspection for cultural heritage: State-of-the-art review.* Journal of Cultural Heritage, 2024. **66**: p. 536-550 DOI: https://doi.org/10.1016/j.culher.2024.01.005.

[16] Bar-Hillel, M., *The base-rate fallacy in probability judgments.* Acta Psychologica, 1980. **44**(3): p. 211-233 DOI: https://doi.org/10.1016/0001-6918(80)90046-3.

[17] Yamashita, R., M. Nishio, R.K.G. Do, and K. Togashi, *Convolutional neural networks: an overview and application in radiology.* Insights into Imaging, 2018. **9**(4): p. 611-629 DOI: https://doi.org/10.1007/s13244-018-0639-9.

[18] Fan, C.-L., *Deep neural networks for automated damage classification in image-based visual data of reinforced concrete structures.* Heliyon, 2024. **10**(19): p. e38104 DOI: https://doi.org/10.1016/j.heliyon.2024.e38104.

[19] Azimi, M., A.D. Eslamlou, and G. Pekcan, *Data-Driven Structural Health Monitoring and Damage Detection through Deep Learning: State-of-the-Art Review.* Sensors, 2020. **20**(10): p. 2778 DOI: https://doi.org/10.3390/s20102778.

[20] Bui-Tien, T., et al., *Damage Detection in Structural Health Monitoring using Hybrid Convolution Neural Network and Recurrent Neural Network.* Frattura ed Integrità Strutturale, 2021. **16**(59): p. 461-470 DOI: https://doi.org/10.3221/IGF-ESIS.59.30.

[21] Deng, Y., Y. Zhao, H. Ju, T.-H. Yi, and A. Li, *Abnormal data detection for structural health monitoring: State-of-the-art review.* Developments in the Built Environment, 2024. **17**: p. 100337 DOI: https://doi.org/10.1016/j.dibe.2024.100337.

[22] Poudel, U.P., G. Fu, and J. Ye, *Structural damage detection using digital video imaging technique and wavelet transformation.* Journal of Sound and Vibration, 2005. **286**(4): p. 869-895 DOI: https://doi.org/10.1016/j.jsv.2004.10.043.

[23] Chen, F. and J.Y. Tsou, *Assessing the effects of convolutional neural network architectural factors on model performance for remote sensing image classification: An in-depth investigation.* International Journal of Applied Earth Observation and Geoinformation, 2022. **112**: p. 102865 DOI: https://doi.org/10.1016/j.jag.2022.102865.

[24] Torzoni, M., L. Rosafalco, A. Manzoni, S. Mariani, and A. Corigliano, *SHM under varying environmental conditions: an approach based on model order reduction and deep learning.* Computers & Structures, 2022. **266**: p. 106790 DOI: https://doi.org/10.1016/j.compstruc.2022.106790.

[25] Alzubaidi, L., et al., *A survey on deep learning tools dealing with data scarcity: definitions, challenges, solutions, tips, and applications.* Journal of Big Data, 2023. **10**(1): p. 46 DOI: https://doi.org/10.1186/s40537-023-00727-2.
[26] Singh, P., N. Singh, K.K. Singh, and A. Singh, *Chapter 5 - Diagnosing of disease using machine learning*, in *Machine Learning and the Internet of Medical Things in Healthcare*, K.K. Singh, et al., Editors. 2021, Academic Press. p. 89-111. DOI https://doi.org/10.1016/B978-0-12-821229-5.00003-3.
[27] Tharwat, A., *Classification assessment methods.* New England Journal of Entrepreneurship, 2020. **17**(1): p. 168-192 DOI: https://doi.org/10.1016/j.aci.2018.08.003.
[28] Stengård, E., P. Juslin, U. Hahn, and R. van den Berg, *On the generality and cognitive basis of base-rate neglect.* Cognition, 2022. **226**: p. 105160 DOI: https://doi.org/10.1016/j.cognition.2022.105160.
[29] Welsh, M.B. and D.J. Navarro, *Seeing is believing: Priors, trust, and base rate neglect.* Organizational Behavior and Human Decision Processes, 2012. **119**(1): p. 1-14 DOI: https://doi.org/10.1016/j.obhdp.2012.04.001.
[30] Autzen, B., *Is the replication crisis a base-rate fallacy?* Theoretical Medicine and Bioethics, 2021. **42**(5): p. 233-243 DOI: https://doi.org/10.1007/s11017-022-09561-8.
[31] Eddy, D.M., *Probabilistic reasoning in clinical medicine: Problems and opportunities*, in *Judgment Under Uncertainty: Heuristics and Biases*, D. Kahneman, P. Slovic, and A. Tversky, Editors. 1982, Cambridge University Press. p. 249--267.
[32] Dahlman, C., *Determining the base rate for guilt.* Law, Probability and Risk, 2017. **17**(1): p. 15-28 DOI: https://doi.org/10.1093/lpr/mgx009.
[33] Webb, M.P.K. and D. Sidebotham, *Bayes' formula: a powerful but counterintuitive tool for medical decision-making.* BJA Educ, 2020. **20**(6): p. 208-213 DOI: https://doi.org/10.1016/j.bjae.2020.03.002.
[34] Egger, S. and E. G., *The vaccinated proportion of people with COVID-19 needs context.* The Lancet, 2022. **399**(10325): p. 627 DOI: https://doi.org/10.1016/S0140-6736(21)02837-3.
[35] Theodoridis, S., *Chapter 2 - Probability and Stochastic Processes*, in *Machine Learning*, S. Theodoridis, Editor. 2015, Academic Press: Oxford. p. 9-51. DOI https://doi.org/10.1016/B978-0-12-801522-3.00002-1.
[36] Webb, G.I., *Bayes Rule*, in *Encyclopedia of Machine Learning*, C. Sammut and G.I. Webb, Editors. 2010, Springer US: Boston, MA. p. 74-75. DOI https://doi.org/10.1007/978-0-387-30164-8_62.
[37] Fenton, N., M. Neil, and D. Berger, *Bayes and the Law.* Annu Rev Stat Appl, 2016. **3**: p. 51-77 DOI: https://doi.org/10.1146/annurev-statistics-041715-033428.
[38] Burciu, Z., *Bayesian methods in reliability of search and rescue action.* Polish Maritime Research, 2010. **17**(4) DOI: https://doi.org/10.2478/v10012-010-0039-7.
[39] O'Kelly, M.E., *Spatial Search and Bayes Theorem: A Commentary on Recent Examples from Aircraft Accidents.* Geographical Analysis, 2023. **55**(3): p. 482-491 DOI: https://doi.org/10.1111/gean.12342.
[40] Rogers, A., G.R. Foxall, and P.H. Morgan, *Building Consumer Understanding by Utilizing a Bayesian Hierarchical Structure within the Behavioral Perspective Model.* Behav Anal, 2017. **40**(2): p. 419-455 DOI: https://doi.org/10.1007/s40614-017-0120-y.